\documentclass{article}

\PassOptionsToPackage{numbers, sort&compress}{natbib}

     \usepackage[preprint]{neurips_2022_ml4ps}

\usepackage[utf8]{inputenc} % allow utf-8 input
\usepackage[T1]{fontenc}    % use 8-bit T1 fonts
\usepackage{hyperref}       % hyperlinks
\usepackage{url}            % simple URL typesetting
\usepackage{booktabs}       % professional-quality tables
\usepackage{amsfonts}       % blackboard math symbols
\usepackage{nicefrac}       % compact symbols for 1/2, etc.
\usepackage{microtype}      % microtypography
\usepackage{xcolor}         % colors
\usepackage{graphicx}
\usepackage{comment}
\usepackage{xcolor}
\usepackage{wrapfig}

\title{Molecular Fingerprints for Robust and Efficient ML-Driven Molecular Generation}

\author{Ruslan N. Tazhigulov, Joshua Schiller, Jacob Oppenheim, Max Winston \\
  EQRx, Inc. \\
  50 Hampshire St., Cambridge, MA 02139, USA \\
  \texttt{\{rtazhigulov, jschiller, joppenheim, mwinston\}@eqrx.com} \\
}

\begin{document}

\maketitle

\begin{abstract}
We propose a novel molecular fingerprint-based variational autoencoder applied for molecular generation on real-world drug molecules. We define more suitable and pharma-relevant baseline metrics and tests, focusing on the generation of diverse, drug-like, novel small molecules and scaffolds. When we apply these molecular generation metrics to our novel model, we observe a substantial improvement in chemical synthetic accessibility ($\Delta\overline{\mbox{SAS}}$ = -0.83) and in computational efficiency up to 5.9x in comparison to an existing state-of-the-art SMILES-based architecture.
\end{abstract}

\section{Introduction}
Machine learning (ML) driven molecular generation is becoming an industry standard among computational and medicinal chemists for exploration of chemical space and drug design. Since the 1980s, the biopharmaceutical industry has used ML methods on two primary tasks in drug development: generating novel molecules and predicting their properties (e.g. quantitative structure activity relationship, QSAR). The deep learning revolution has enabled these tasks to move from disparate strategies, to models based on the embedding of chemical structures in a high dimensional space, either implicitly or explicitly. We focus on the first task here, as the second task remains limited in applicability by the size of training datasets outside a handful of well-studied molecular sets and properties.

Representation of the chemical structures of the molecules can be split into two major classes: 1D string-based representations, and 2D/3D molecular graphs.
While 2D/3D graphical representations can directly capture the connectivity between atoms, graphs are provably more difficult and slow to generate than 1D string sequences as graph generation is known to be an NP-hard problem \cite{GenModels:WIREsReview,Molassembler:jcim:2020}.
The most common type of 1D string representation of molecules is SMILES (Simplified Molecular Input Line Entry System) \cite{SMILES:1988}. This simplistic string-based representation transforms a molecule into a sequence of characters based on the set of predefined atom ordering rules. Using 1D strings for molecular representation has proven beneficial, enabling the reapplication of existing neural network architectures that have been previously developed for natural language processing (NLP) tasks.

In principle, there is hardly any limit on representation of the molecules with SMILES strings. While one molecule can map to its multiple equivalent SMILES representations, typically, only one canonical SMILES representation is used. \citet{Arus:randomizedSMILES:2019} showed that augmenting training data with randomized versions of SMILES strings leads to generalization to larger drug-like chemical spaces than solely canonical SMILES (single representation), and increases model performance on generative tasks. A model trained with randomized SMILES was able to generate at least double the amount of unique molecules compared to models based on canonical SMILES. 

Although several various model architectures have been developed (VAEs \cite{Bombarelli:ChemVAE:2018, Bayer:VAE:2019}, cRNN \cite{Kotsias:cRNN:2020}, GAN \cite{ORGAN-SMILES:Aspuru-Guzik:arxiv:2017}, etc), a lack of meaningful standardized metrics and benchmarks for molecular generation makes it difficult to evaluate and compare them for their actual application in the drug discovery process. For example, despite the increasing prevalence of molecular generation in the pharmaceutical industry, performance in the literature largely continues to be measured by on-the-fly metrics such as reconstruction accuracy. While these on-the-fly metrics are important in the training and testing processes, ultimately metrics capturing computationally-efficient generation of novel chemical matter are more relevant.
Given the utility of structural chemical variation in modern drug discovery,
capturing unique scaffolds \cite{BemisMurcko:1996} and structural similarity of generated molecules within metrics is vital for the field to advance.

Building on these results, we investigated the use of molecular fingerprints for generating novel chemical matter instead of using SMILES as an input. Unlike SMILES representations, any molecular structure can be deterministically mapped to its fingerprint. Previously it had been assumed that molecular fingerprints are completely uninvertible due to the complex rules of construction and the hash mapping used in implementation. However, \citet{Clevert:Neuraldecipher-Bayer:2020} have shown that circular extended connectivity fingerprints (ECFP) are at least partially invertible. 
Since it is likely that circular molecular fingerprints have a better inductive bias compared to SMILES by construction, comparative analysis of the molecular generation efficacy between circular fingerprints and SMILES was a practical investigation.
Thus, the invertibility of circular molecular fingerprints presented an opportunity for their use in molecular generators since they represent ideal encoders: they are fast to compute, contain a high amount of information, and are related to underlying molecular graphs by design. Given all of these advantages, we sought to build a novel fingerprint-based architecture that could outperform existing state-of-the-art SMILES-based molecular generators. 

Here, we set out to benchmark and synthesize existing literature approaches, build best-in-class models, and propose a novel molecular fingerprint-based variational autoencoder (VAE) for molecular generation.~In addition to the common on-the-fly metrics used in the literature, we benchmark the implemented models with the computational efficiency of the molecular generation process, determined by the number of structurally unique molecules that a model has generated. Furthermore, in an attempt to highlight the most relevant chemical novelty, we focus on characterizing novel chemical matter with respect to chemical distance from reference target molecules. We thus attempt to define pharma-relevant benchmarks and tests for molecular generation focused on the generation of drug-like novel small molecules and scaffolds.

\section{Methods}

\subsection{Molecular ChemVAE: SMILES-based variational autoencoder} 
In the present work, we trained a molecular ChemVAE model that is descended from the one presented by \citet{Bombarelli:ChemVAE:2018}. 
We applied numerous architecture and training optimizations suggested in the ML molecular generation literature. Among them, we benchmarked multiple accuracy metrics (Table~\ref{tab:chemvae_fpvae_datasets}), while still focusing on molecular generation capabilities of the models.
Note that while the original ChemVAE was trained on ZINC and QM9 datasets \cite{Bombarelli:ChemVAE:2018}, we opted for ChEMBL \cite{ChEMBL:2018}, a manually curated database of bioactive molecules with drug-like properties. Our cleaned-up dataset contained 1.83M of unique and valid, from SMILES syntax perspective, molecules. 90 percent of the molecules were used for training, with the 10 percent allocated for testing. For ease of computation, we put a cap of 200 on the maximum length of the SMILES strings.
We also implemented the approach discussed by \citet{Arus:randomizedSMILES:2019}, randomizing the original canonical SMILES strings on the fly. This procedure results in a different SMILES string for the same molecule in each epoch. More details on model parameters and hyperparameters can be found below.

\subsubsection{Data preparation} 
We used ChEMBL-28 database \cite{ChEMBL:2018} (named as ChEMBL-1.83M in this work).
For our early exploration, we also used a subset of ChEMBL-28 database, namely ChEMBL-567k.
For ChEMBL-567k, we only included the active compounds with IC50/EC50 values $\geq$ 6, similar to \citet{Schneider:RNNgen:2018}, and we restricted the range for the length of included SMILES strings between 10 and 100 for computational purposes. Note that \citet{Bombarelli:ChemVAE:2018} used a subset of ZINC dataset (namely ZINC-249k in Table~\ref{tab:chemvae_fpvae_datasets}) for training of the original ChemVAE model.

We canonicalized the SMILES strings using RDKit package. We treated compounds as racemates, i.e. any information about stereochemistry (represented by "@", "/", and "$\backslash$ $\backslash$" tokens) was removed. We also removed all ionic non-bonded components of the original SMILES strings (represented by "."). 
For ChEMBL-1.83M, we restricted the maximal length to 215 to account for equivalent randomized SMILES representations that are longer than their original canonical SMILES counterparts. The shorter strings were padded with spaces to this same length.

\subsubsection{SMILES tokenizer} 
On top of multiple equivalent representations, SMILES strings can be tokenized in various ways, and more advanced tokenizers have been emerging. In the present work, we relied on the atomwise tokenizer from SMILES Pair Encoding (SPE) package \cite{Fourches:atomwise:2021}. SPE package is licensed under Apache-2.0 license. SPE aims to augment the widely used atom-level SMILES tokenization by adding chemically explainable and human-readable SMILES substrings as tokens. Case studies showed that SPE could demonstrate superior performances on both molecular generation tasks with regard to novelty, diversity, and ability to resemble the training set distribution \cite{Fourches:atomwise:2021}. 
Therefore, our encoding used 193 tokens, which included auxiliary tokens obtained during the randomized SMILES procedure. 

\subsubsection{Model} 
\begin{wrapfigure}{r}{0.6\textwidth}
  \centering
  \vspace{-10pt}
  \includegraphics[width=0.60\textwidth]{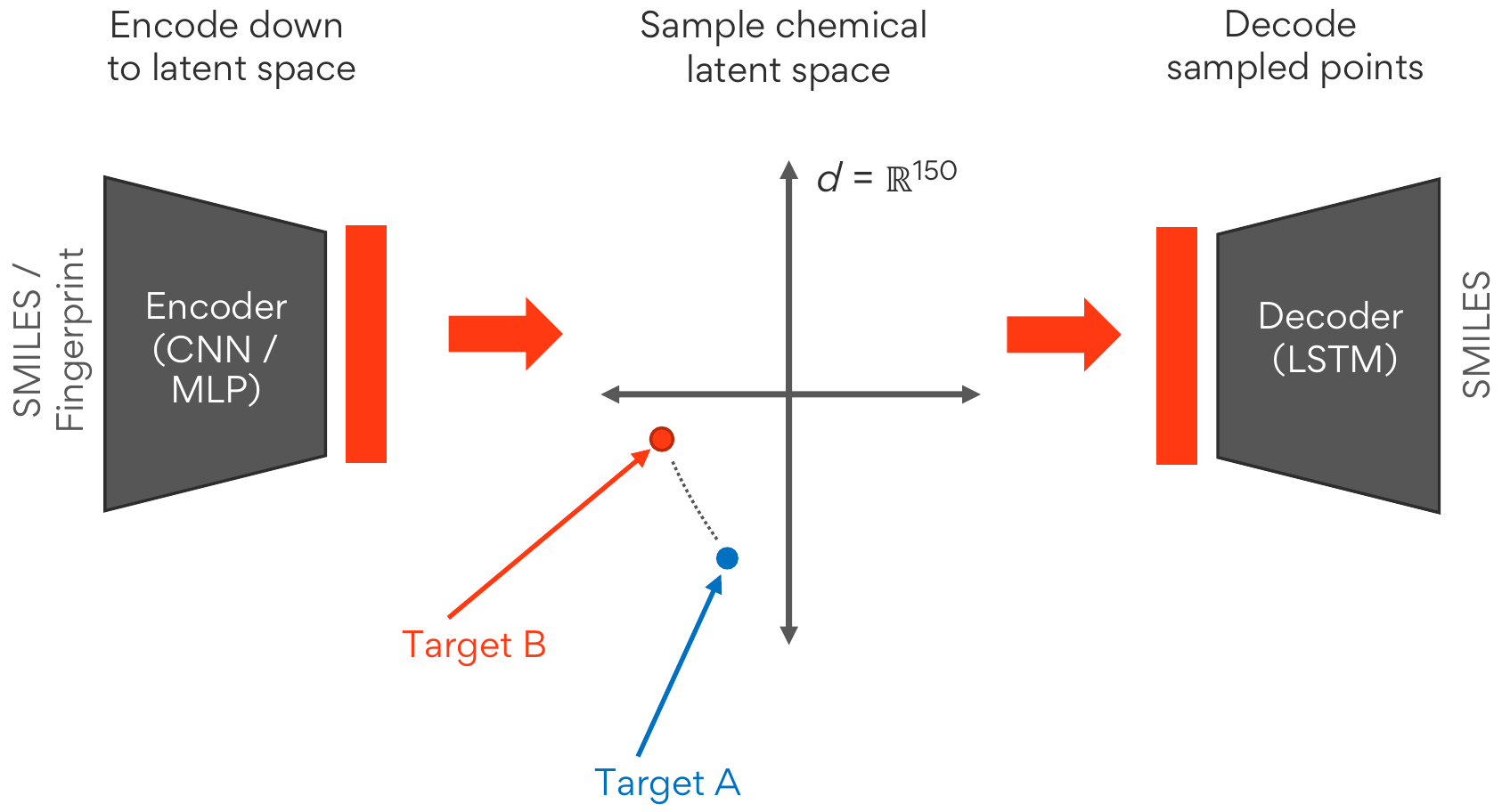}
  \caption{A schematic representation of variational autoencoders trained and applied for molecular generation in the present work.}
  \vspace{-9pt}
  \label{fig:models}
\end{wrapfigure}
The VAE deep network structure of our molecular ChemVAE model was as follows (Fig.~\ref{fig:models}): the encoder used three 1D convolutional layers of 9, 9, 10 convolution kernels with the kernel sizes of 9, 9, 11, respectively, followed by two fully connected layers with dimensions of 435 and 150. The latter one corresponds to the dimension of the chemical latent space vectors, with the vector values bounded between -1 and 1 by applying tanh activation function. Batch normalization was enforced at the encoder level after each convolution layer.
The decoder included three stacked layers of long short-term memory (LSTM) networks \cite{LSTM:1997} with the hidden dimension of 512.
The last layers included a fully-connected layer of the vocabulary dimension, followed by the softmax layer with the argmax being applied.
The variational (KL) loss was annealed according to sigmoid schedule after 20 epochs, running for a total 100 epochs, and He-initialization was applied to all fully-connected layers. The batch size was set to 256. AdamW optimizer \cite{AdamW:2017} with the learning rate of 0.0001 and weight decay of 0.001 was used for backpropagation, and we also used a learning rate scheduler: the learning rate was decreased by a factor of 0.7 if the token-by-token reconstruction accuracy on the testing set was not improved within the next 5 epochs. Both cross-enthropy and KL losses were averaged over all tokens that combined into a total loss. The best performance model based on the highest token-by-token reconstruction accuracy for the testing set was chosen further for molecular generation tasks.

\subsection{Molecular FpVAE: fingerprint / SMILES-based variational autoencoder}
Our Fp / SMILES-based variational autoencoder (FpVAE) also used an entire ChEMBL dataset.
We obtained circular Morgan fingerprints (non-count version, nBits = 4096, radius = 2), an analogous open-source version of the ECFP fingerprints, for each molecule in the dataset. We have not observed any significant benefits from increasing the number of bits or radius parameter, as \citet{Clevert:Neuraldecipher-Bayer:2020} also observed and indicated for the ECFP fingerprints in their work.
The same 90/10 split was used between the training and testing sets. 
We opted for the canonical SMILES representation of the molecules for training of the FpVAE model. More details on model parameters and hyperparameters can be found below.

\begin{table}[!htbp]
  \caption{Token-by-token, molecular reconstruction, and Tanimoto accuracy (computed between corresponding Morgan fingerprints) metrics obtained for the test sets of molecular ChemVAE and FpVAE models trained with different datasets}
  \label{tab:chemvae_fpvae_datasets}
  \centering
  \begin{tabular}{lllll}
    \toprule
    Architecture & Dataset & Token-by-token & Molecule & Tanimoto \\
    \midrule
    ChemVAE & ZINC-249k & 0.972 & - & - \\
    ChemVAE & ChEMBL-567k & 0.978 & - & - \\
    ChemVAE & ChEMBL-1.83M & 0.996 & 0.688 & 0.816 \\
    FpVAE & ChEMBL-567k & 0.878 & - & - \\
    FpVAE & ChEMBL-1.83M & 0.948 & 0.360 & 0.508 \\
    \bottomrule
  \end{tabular}
\end{table}

\subsubsection{SMILES tokenizer} 
For tokenization of the output, we used the same atomwise tokenizer \cite{Fourches:atomwise:2021} that we also applied for the molecular ChemVAE model dataset. Encoding on the virtually entire ChEMBL dataset (canonical SMILES) used 184 tokens.

\subsubsection{Model}
The encoder part contained five fully-connected layers with the dimensions of 2048, 1024, 768, and 512, respectively, with the last layer defining the dimension of the latent space, 150 (Fig.~\ref{fig:models}). Our early results also demonstrated no gain in performance when the dropout for fully-connected layers in the encoder was applied. We set the dropout rates to 0.05, 0.10, 0.15, and 0.20, and similarly to what \citet{Clevert:Neuraldecipher-Bayer:2020} indicated in their work, we have not observed any further improvements from using dropout regularization (Table~\ref{tab:fpvae_dropoutrate}). 
The decoder part contained three stacked layers of LSTM networks \cite{LSTM:1997} with the hidden dimension of 512. The last layers included a fully-connected layer of the vocabulary dimension, followed by the softmax layer with the argmax being applied. The variational (KL) loss was annealed according to the same sigmoid schedule after 20 epochs, running for a total 200 epochs, and He-initialization was applied to all fully-connected layers. The batch size was set to 256. AdamW optimizer \cite{AdamW:2017} with the learning rate of 0.0005 and weight decay of 0.005 was used for backpropagation, and we also used a learning rate scheduler: the learning rate was decreased by a factor of 0.7 if the token-by-token reconstruction accuracy on the testing set was not improved within the next 5 epochs. Both cross-enthropy and KL losses were averaged over all tokens that combined into a total loss. The best performance model based on the highest token-by-token reconstruction accuracy for the testing set was chosen further for molecular generation tasks.

\begin{table}[!htbp]
  \caption{Token-by-token, molecular reconstruction, and Tanimoto accuracy (computed between corresponding Morgan fingerprints) metrics obtained for the test sets of FpVAE models (ChEMBL-567k dataset) trained with different dropout rates in fully connected layers. Numbers in the parentheses are computed for the decoded valid SMILES only}
  \label{tab:fpvae_dropoutrate}
  \centering
  \begin{tabular}{llll}
    \toprule
    Dropout rate & Token-by-token & Molecule & Tanimoto \\
    \midrule
    0.00 & 0.878 & 0.141 (0.382) & 0.293 (0.794) \\
    0.05 & 0.873 & 0.092 (0.299) & 0.232 (0.753) \\
    0.10 & 0.862 & 0.055 (0.228) & 0.171 (0.710) \\
    0.15 & 0.851 & 0.034 (0.187) & 0.126 (0.692) \\
    0.20 & 0.836 & 0.014 (0.116) & 0.075 (0.620) \\
    \bottomrule
  \end{tabular}
\end{table}

\subsection{MG-Bridge: SLERP-based navigation in chemical latent space for molecular generation} 
A high-dimensional chemical latent space generated by the ML models tends to be substantially sparse. 
Effective sampling of the sparse chemical latent space is becoming a cornerstone of ML-driven molecular generation process. For example, one can imagine that perturbing a latent space vector with a Gaussian noise can potentially produce a new molecule. 
However, the optimal perturbation noise that allows one to explore the latent space efficiently and produce a substantial number of unique and novel molecules per computational time remains unclear. Moreover, drug discovery researchers typically focus on a few (typically a pair) of active compounds, not just a single compound. 
\begin{wrapfigure}{l}{0.3\textwidth}
  \centering
  \includegraphics[width=0.3\textwidth]{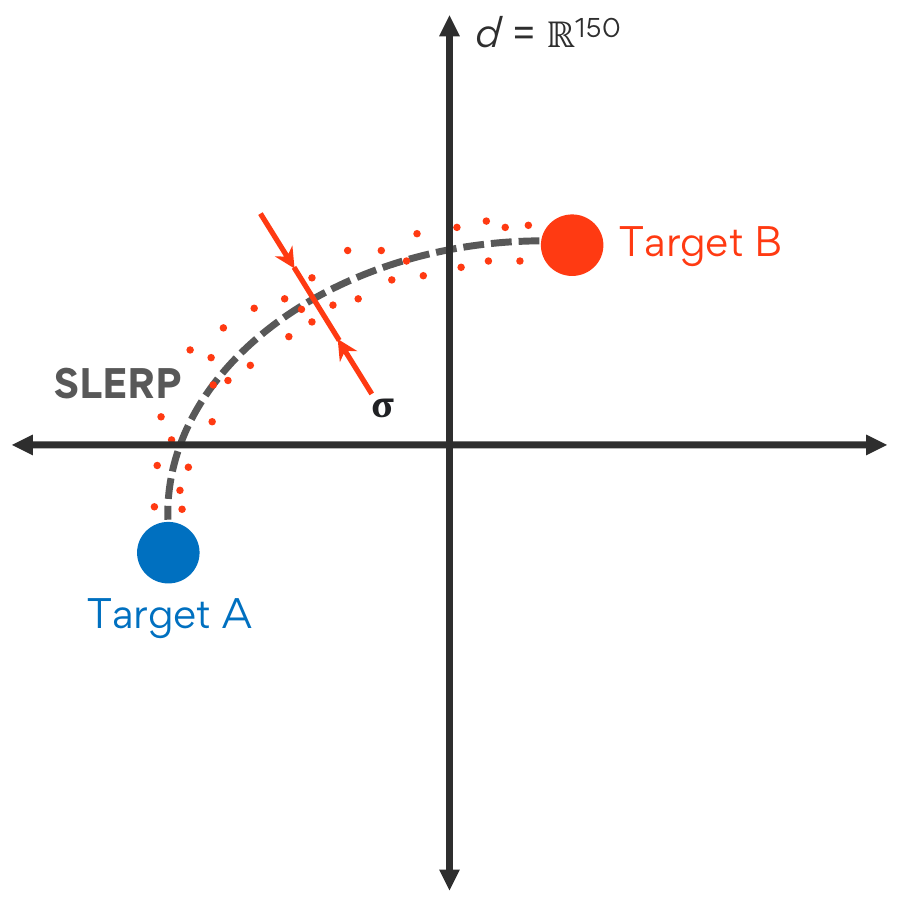}
  \vspace{-15pt}
  \caption{Molecular generation bridge (MG-Bridge).}
  \label{fig:slerp}
  \vspace{-10pt}
\end{wrapfigure}
Hence, effective techniques that allow one to explore the latent space between multiple active compounds are needed.
\citet{Bombarelli:ChemVAE:2018} introduced the idea of sampling along the trajectories between a pair of reference molecules in the latent space to generate novel molecules based on linear or spherical linear interpolation (LERP/SLERP). 
Here, we introduce a molecular generation procedure named as MG-Bridge (Molecular Generation Bridge). 
The procedure combines SLERP and perturbation of latent space vectors with a Gaussian noise ($\sigma$) into one method (Fig.~\ref{fig:slerp}, more details can be found below. MG-Bridge enables efficient molecular generation and exploration of the chemical latent space between a pair of active reference compounds.

\textbf{MG-Bridge: details on chemical latent space sampling.} The coarse-grained MG-Bridge runs were performed as follows: 100 grid points along the SLERP trajectory between molecular target A and target B were perturbed 100 times, resulting in 10000 latent space vectors that were then decoded into SMILES. For each reference molecular pair, we scanned across multiple parameters of the perturbation noise ($\sigma$) to obtain an optimal perturbation noise per pair (Fig.~\ref{fig:scan}). Once the optimal noise $\sigma$ was obtained, we used the optimized value for production MG-Bridge runs: 100 grid points along the SLERP trajectory were perturbed 5000 times, resulting in 500000 latent space vectors that were then decoded into SMILES strings.

\section{Results and Discussion}

\begin{wraptable}{r}{5.5cm}
\vspace{-15pt}
  \caption{Reference pair molecules used for benchmarking of VAE models}
  \label{tab:reference_compounds}
  \centering
  \begin{tabular}{lll}
    \toprule
    Class & Target A & Target B \\
    \midrule
    NSAID & Ibuprofen & Naproxen \\
    EGFR & Gefitinib & Erlotinib \\
    VEGFR & Pazopanib & Sunitinib \\
    PI3K & Alpelisib & Inavolisib \\
    \bottomrule
  \end{tabular}
  \vspace{-15pt}
\end{wraptable}
We benchmarked the molecular generation performance of both existing molecular ChemVAE and a new FpVAE models for four reference pairs of drug molecules. Outside of NSAIDs, they reflect recent and current areas of drug development spanning different levels of molecular complexity (Table~\ref{tab:reference_compounds}). We applied chemoinformatic filters to the obtained molecular generation sets based on quantitative estimate of drug-likeliness (QED) \cite{QEDscore:2012}, synthetic accessibility score (SAS) \cite{SAS:2009}, and novel generic Bemis-Murcko (BM) scaffolds \cite{BemisMurcko:1996}. We incorporated these filters in our benchmarks.

\begin{wrapfigure}{l}{0.3\textwidth}
  \vspace{-10pt}
  \centering
  \includegraphics[width=0.3\textwidth]{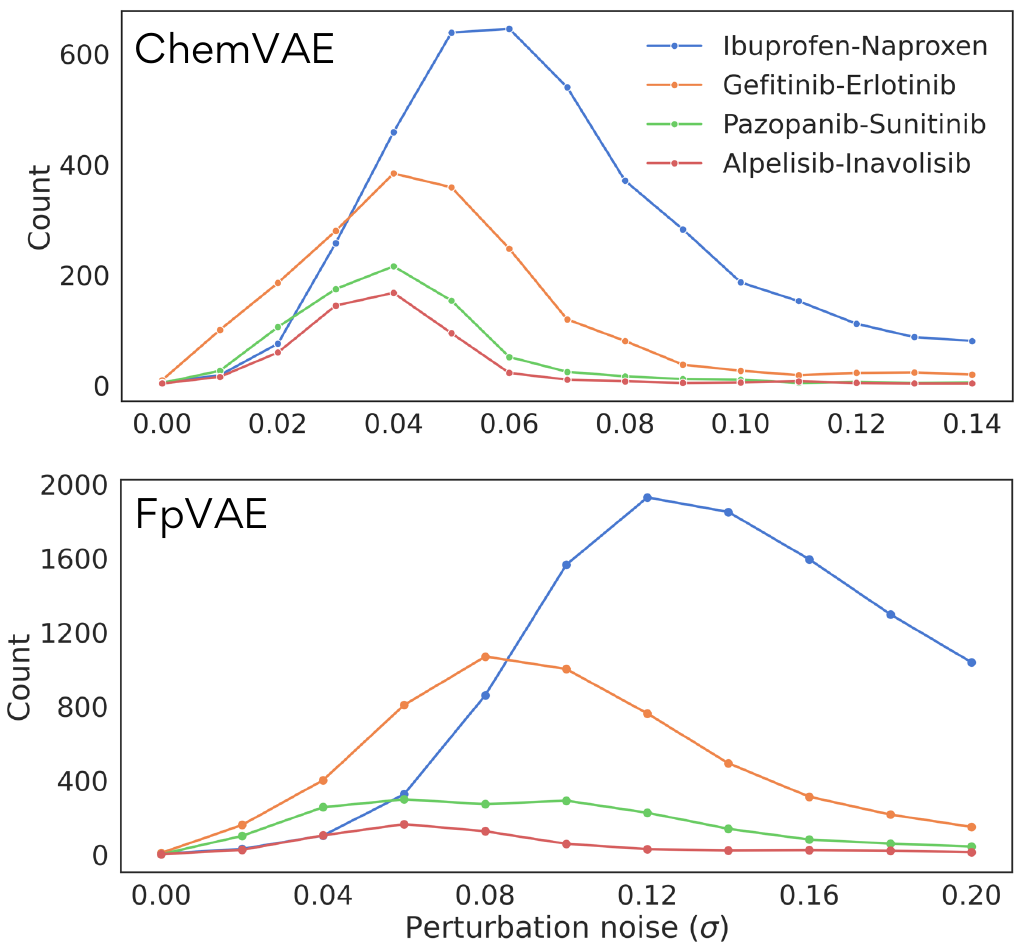}
  \vspace{-15pt}
  \caption{MG-Bridge scan for optimal perturbation noise.}
  \vspace{-15pt}
  \label{fig:scan}
\end{wrapfigure}
Our results indicate the importance of introducing perturbation in SLERP and using MG-Bridge. We show that SLERP has to be enhanced for the real-world applications, because a pure application of SLERP led to a handful (O(10)) of unique and novel molecules (Fig.~\ref{fig:scan}, $\sigma$ = 0.0) across all four reference pairs. 
Once perturbation is applied, it becomes clear that the effect of perturbation and its size varies for different molecular pairs. The results also indicate different levels of sparsity in the chemical latent space across all pairs and various model architectures. One should note that the novelty--a metric that is typically used to account for molecules not present in the original training set--of molecules generated by the MG-Bridge is >99\% for the obtained valid (from SMILES perspective) and unique molecules.

\begin{wrapfigure}{r}{0.45\textwidth}
  \vspace{-1pt}
  \centering
  \includegraphics[width=0.45\textwidth]{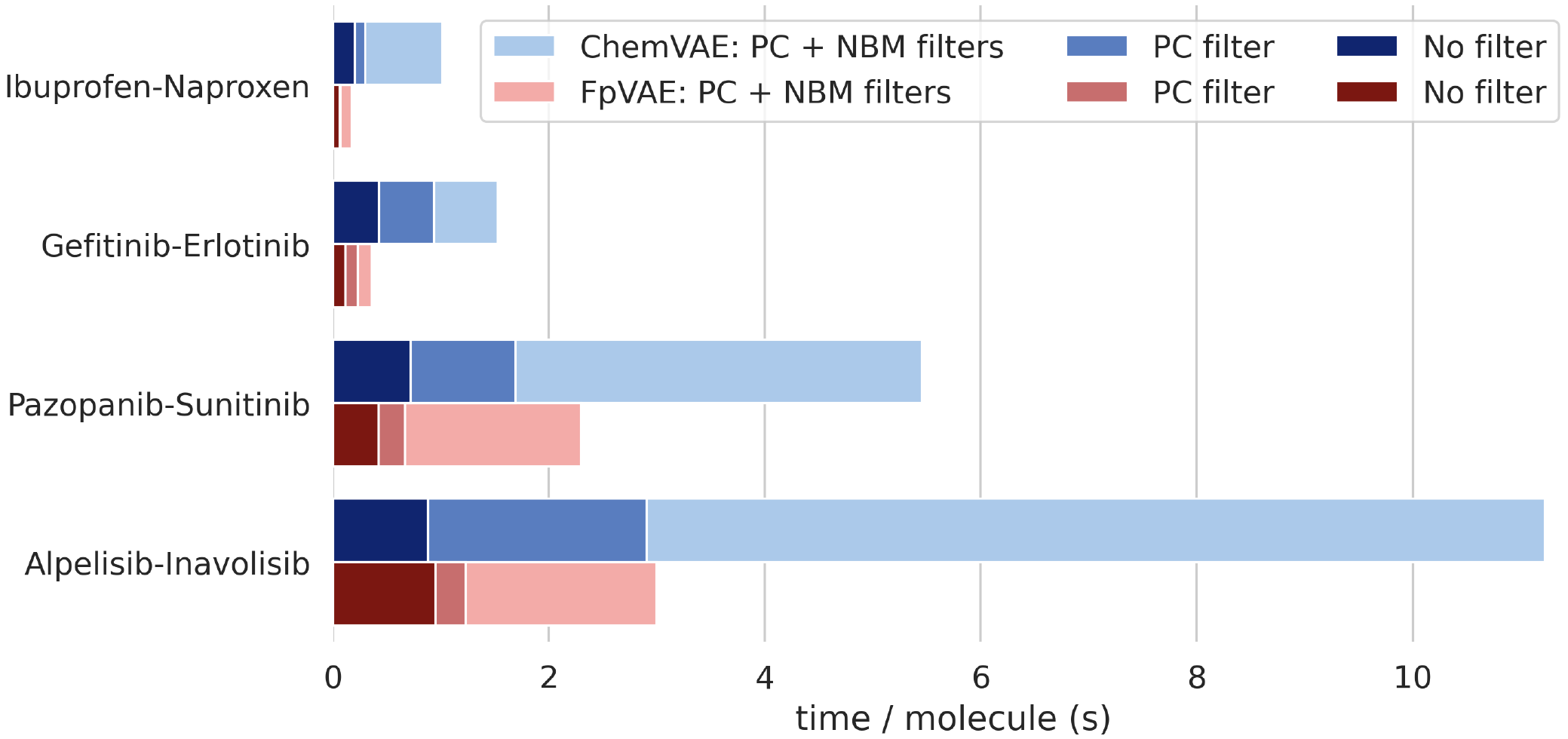}
  \caption{Efficiency of molecular generation and key performance metrics across four reference target pairs.}
  \label{fig:kpis}
  \vspace{-10pt}
\end{wrapfigure}

Furthermore, our results show that the FpVAE model significantly outperforms the ChemVAE model in molecular generation computational efficiency$\footnote{NVIDIA Tesla T4, 16 CPUs, 60GB RAM}$ by as large as a factor of 5.9, even more notably when physicochemical property (PC filter: QED $\geq$ 0.4, SAS $\leq$ 4.0) and novel BM (NBM) scaffold filters are applied (Fig.~\ref{fig:kpis}). The PC filter is aimed to filter out the molecules that are not drug-like or hard to synthesize. The NBM filter is aimed to keep the molecules with the generic BM scaffolds not belonging to the original reference molecules, and therefore, it aims to incorporate the chemical structural diversity.

\begin{figure}[htbp]
  \centering
  \vspace{-5pt}
  \includegraphics[width=0.78\textwidth]{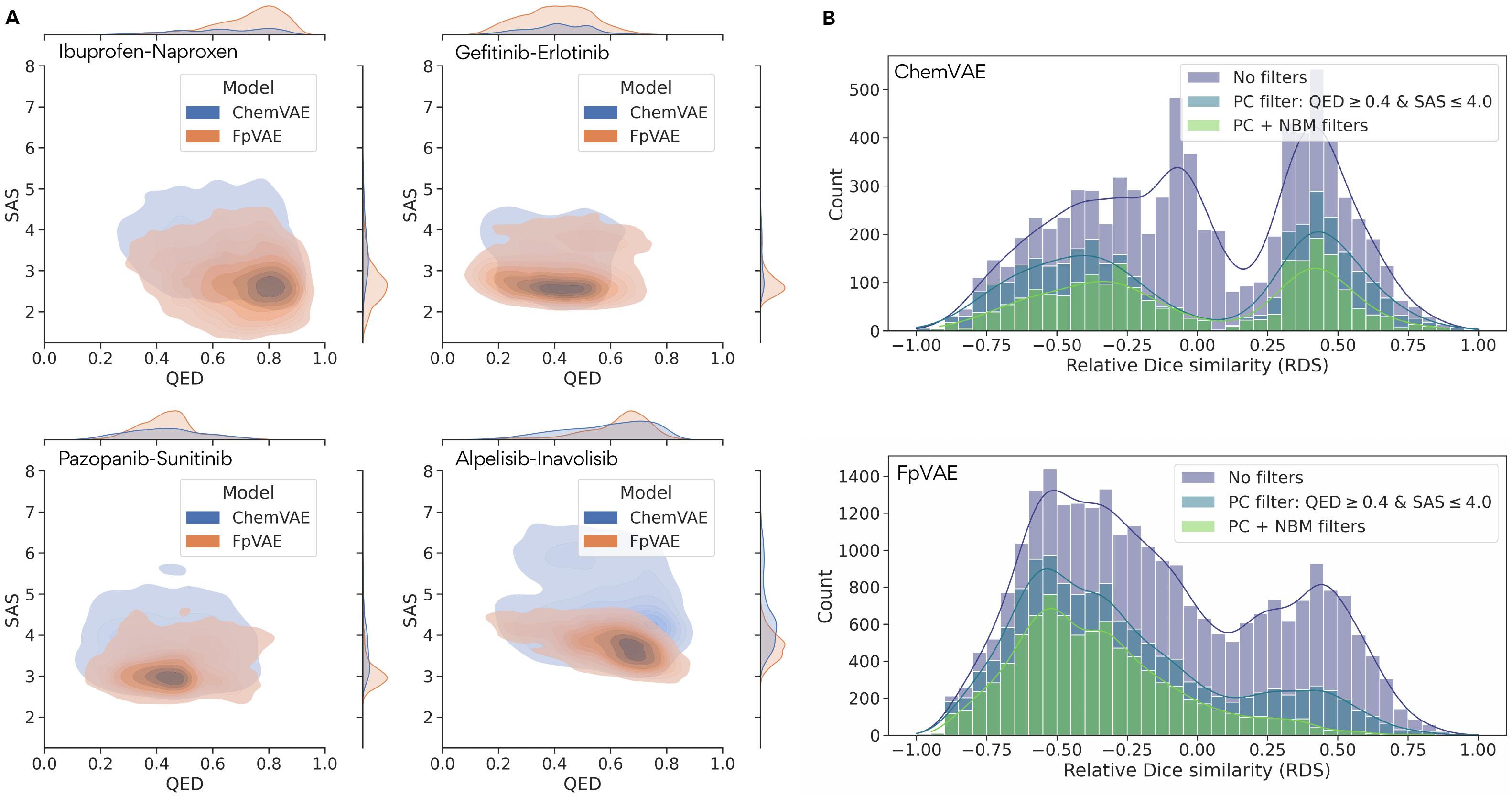}
  \caption{(A) QED-SAS properties computed for the sets obtained with four reference pair molecules. The larger the QED and the smaller the SAS scores (bottom right) indicates better results.
  (B) RDS computed for the molecular generation sets obtained for EGFR inhibitors (Gefitinib-Erlotinib pair).}
  \label{fig:qed-sas-contour_dice}
  \vspace{-5pt}
\end{figure}

Moreover, by more rigorously examining the physicochemical properties of generated molecules, we demonstrate that the molecular FpVAE outperforms the molecular ChemVAE model. 
FpVAE is able to generate more sensible, drug-like and synthesizable molecules across all four reference pairs (Fig.~\ref{fig:qed-sas-contour_dice}A).
These results imply that the FpVAE learns chemical rules substantially more efficiently than the SMILES-based ChemVAE. Furthermore, they also indicate that circular molecular fingerprints have a better inductive bias compared to SMILES, demonstrating the efficacy of fingerprints for molecular generation.
Since the same number of latent space vectors is sampled (with the optimal perturbation noise) for both FpVAE and ChemVAE models, we also show that significantly larger number of unique and valid molecules are generated with the FpVAE model (Fig.~\ref{fig:qed-sas-contour_dice}A).

We also evaluated structural diversity in the molecular generation sets by computing the relative Dice similarity, $RDS_i = \frac{d_{Bi} - d_{Ai}}{1 - d_{AB}}$, where $d$ is Dice similarity between two Morgan fingerprints corresponding to two different molecules. RDS $\in$ [-1.0, 1.0], where the negative values indicate that a generated molecule is chemically similar to target A, while the positive values indicate that a molecule is chemically similar to target B. The results in Fig.~\ref{fig:qed-sas-contour_dice}B indicate that the molecular FpVAE model is able to generate a substantially larger number of unique and valid molecules compared to ChemVAE for the optimal perturbation noise and with the same number of latent space vectors provided for decoding. Moreover, molecules generated by FpVAE are more drug-like and more likely synthesizable, i.e. more of them pass the PC filter. 
Furthermore, when it comes to chemical diversity in the generated sets, the FpVAE model is able to generate significantly larger number of molecules with NBM scaffolds. One should note that the most chemically differentiated and diverse molecules lay in the interval of RDS $\in$ (-0.3, 0.3), and the FpVAE model generates a substantially larger number of molecules in this interval. 
These results also illustrate that the chemical latent space, generated by both ChemVAE and FpVAE models, is not really homogeneous and the sparsity varies, e.g. it is easier to generate more chemically sensible and diverse molecules in the vicinity of gefitinib vs. erlotinib. The sparsity and compactness of the chemical latent spaces as well as using circular molecular fingerprints in combination with SMILES as an input for the encoder will be addressed in the future work.

\section{Conclusion}
In the present work, we proposed and discussed the performance of a novel variational autoencoder based on the molecular fingerprints (FpVAE). 
Importantly, we also proposed a set of baseline pharma-relevant benchmarks for comparison of molecular generation performance across various model architectures, methods, molecules, and filters that can be applied for unbiased comparative analysis. These benchmarks put a focus on the number of drug-like, synthesizable, and structurally diverse molecules generated per computational time.
We showed that the FpVAE model is able to generate molecules more efficiently compared to the purely SMILES-based ChemVAE model. We also highlighted that the FpVAE generation sets obtained for various reference drug molecule pairs representing different classes outperform the ones obtained with the purely SMILES-based model in chemical synthetic accessibility and structural diversity.

{
\small
\bibliography{biblist}

% Generated by IEEEtranN.bst, version: 1.14 (2015/08/26)
\begin{thebibliography}{17}
\providecommand{\natexlab}[1]{#1}
\providecommand{\url}[1]{#1}
\csname url@samestyle\endcsname
\providecommand{\newblock}{\relax}
\providecommand{\bibinfo}[2]{#2}
\providecommand{\BIBentrySTDinterwordspacing}{\spaceskip=0pt\relax}
\providecommand{\BIBentryALTinterwordstretchfactor}{4}
\providecommand{\BIBentryALTinterwordspacing}{\spaceskip=\fontdimen2\font plus
\BIBentryALTinterwordstretchfactor\fontdimen3\font minus
  \fontdimen4\font\relax}
\providecommand{\BIBforeignlanguage}[2]{{%
\expandafter\ifx\csname l@#1\endcsname\relax
\typeout{** WARNING: IEEEtranN.bst: No hyphenation pattern has been}%
\typeout{** loaded for the language `#1'. Using the pattern for}%
\typeout{** the default language instead.}%
\else
\language=\csname l@#1\endcsname
\fi
#2}}
\providecommand{\BIBdecl}{\relax}
\BIBdecl

\bibitem[Bilodeau et~al.(2022)Bilodeau, Jin, Jaakkola, Barzilay, and
  Jensen]{GenModels:WIREsReview}
C.~Bilodeau, W.~Jin, T.~Jaakkola, R.~Barzilay, and K.~F. Jensen, ``Generative
  models for molecular discovery: Recent advances and challenges,'' \emph{WIREs
  Computational Molecular Science}, p. e1608, 2022.

\bibitem[Sobez and Reiher(2020)]{Molassembler:jcim:2020}
J.-G. Sobez and M.~Reiher, ``Molassembler: Molecular graph construction,
  modification, and conformer generation for inorganic and organic molecules,''
  \emph{Journal of Chemical Information and Modeling}, vol.~60, no.~8, pp.
  3884--3900, 2020.

\bibitem[Weininger(1988)]{SMILES:1988}
D.~Weininger, ``Smiles, a chemical language and information system. 1.
  introduction to methodology and encoding rules,'' \emph{Journal of Chemical
  Information and Computer Sciences}, vol.~28, no.~1, pp. 31--36, 1988.

\bibitem[Ar{\'u}s-Pous et~al.(2019)Ar{\'u}s-Pous, Johansson, Prykhodko,
  Bjerrum, Tyrchan, Reymond, Chen, and Engkvist]{Arus:randomizedSMILES:2019}
J.~Ar{\'u}s-Pous, S.~V. Johansson, O.~Prykhodko, E.~J. Bjerrum, C.~Tyrchan,
  J.-L. Reymond, H.~Chen, and O.~Engkvist, ``Randomized smiles strings improve
  the quality of molecular generative models,'' \emph{Journal of
  cheminformatics}, vol.~11, no.~1, pp. 1--13, 2019.

\bibitem[Gómez-Bombarelli et~al.(2018)Gómez-Bombarelli, Wei, Duvenaud,
  Hernández-Lobato, Sánchez-Lengeling, Sheberla, Aguilera-Iparraguirre,
  Hirzel, Adams, and Aspuru-Guzik]{Bombarelli:ChemVAE:2018}
R.~Gómez-Bombarelli, J.~N. Wei, D.~Duvenaud, J.~M. Hernández-Lobato,
  B.~Sánchez-Lengeling, D.~Sheberla, J.~Aguilera-Iparraguirre, T.~D. Hirzel,
  R.~P. Adams, and A.~Aspuru-Guzik, ``Automatic chemical design using a
  data-driven continuous representation of molecules,'' \emph{ACS Central
  Science}, vol.~4, no.~2, pp. 268--276, 2018.

\bibitem[Winter et~al.(2019)Winter, Montanari, Noé, and
  Clevert]{Bayer:VAE:2019}
R.~Winter, F.~Montanari, F.~Noé, and D.-A. Clevert, ``Learning continuous and
  data-driven molecular descriptors by translating equivalent chemical
  representations,'' \emph{Chem. Sci.}, vol.~10, pp. 1692--1701, 2019.

\bibitem[Kotsias et~al.(2020)Kotsias, Ar{\'u}s-Pous, Chen, Engkvist, Tyrchan,
  and Bjerrum]{Kotsias:cRNN:2020}
P.-C. Kotsias, J.~Ar{\'u}s-Pous, H.~Chen, O.~Engkvist, C.~Tyrchan, and E.~J.
  Bjerrum, ``Direct steering of de novo molecular generation with descriptor
  conditional recurrent neural networks,'' \emph{Nature Machine Intelligence},
  vol.~2, no.~5, pp. 254--265, 2020.

\bibitem[Guimaraes et~al.(2017)Guimaraes, Sanchez-Lengeling, Outeiral, Farias,
  and Aspuru-Guzik]{ORGAN-SMILES:Aspuru-Guzik:arxiv:2017}
\BIBentryALTinterwordspacing
G.~L. Guimaraes, B.~Sanchez-Lengeling, C.~Outeiral, P.~L.~C. Farias, and
  A.~Aspuru-Guzik, ``Objective-reinforced generative adversarial networks
  ({ORGAN}) for sequence generation models,'' 2017. [Online]. Available:
  \url{https://arxiv.org/abs/1705.10843}
\BIBentrySTDinterwordspacing

\bibitem[Bemis and Murcko(1996)]{BemisMurcko:1996}
G.~W. Bemis and M.~A. Murcko, ``The properties of known drugs. 1. molecular
  frameworks,'' \emph{Journal of Medicinal Chemistry}, vol.~39, no.~15, pp.
  2887--2893, 1996.

\bibitem[Le et~al.(2020)Le, Winter, Noé, and
  Clevert]{Clevert:Neuraldecipher-Bayer:2020}
T.~Le, R.~Winter, F.~Noé, and D.-A. Clevert, ``Neuraldecipher –
  reverse-engineering extended-connectivity fingerprints ({ECFPs}) to their
  molecular structures,'' \emph{Chem. Sci.}, vol.~11, pp. 10\,378--10\,389,
  2020.

\bibitem[Mendez et~al.(2018)Mendez, Gaulton, Bento, Chambers, De Veij, Félix,
  Magariños, Mosquera, Mutowo, Nowotka, Gordillo-Marañón, Hunter, Junco,
  Mugumbate, Rodriguez-Lopez, Atkinson, Bosc, Radoux, Segura-Cabrera, Hersey,
  and Leach]{ChEMBL:2018}
D.~Mendez, A.~Gaulton, A.~P. Bento, J.~Chambers, M.~De Veij, E.~Félix,
  M.~Magariños, J.~Mosquera, P.~Mutowo, M.~Nowotka, M.~Gordillo-Marañón,
  F.~Hunter, L.~Junco, G.~Mugumbate, M.~Rodriguez-Lopez, F.~Atkinson, N.~Bosc,
  C.~Radoux, A.~Segura-Cabrera, A.~Hersey, and A.~Leach, ``{ChEMBL: towards
  direct deposition of bioassay data},'' \emph{Nucleic Acids Research},
  vol.~47, no.~D1, pp. D930--D940, 11 2018.

\bibitem[Gupta et~al.(2018)Gupta, Müller, Huisman, Fuchs, Schneider, and
  Schneider]{Schneider:RNNgen:2018}
A.~Gupta, A.~T. Müller, B.~J.~H. Huisman, J.~A. Fuchs, P.~Schneider, and
  G.~Schneider, ``Generative recurrent networks for de novo drug design,''
  \emph{Molecular Informatics}, vol.~37, no. 1-2, p. 1700111, 2018.

\bibitem[Li and Fourches(2021)]{Fourches:atomwise:2021}
X.~Li and D.~Fourches, ``Smiles pair encoding: A data-driven substructure
  tokenization algorithm for deep learning,'' \emph{Journal of Chemical
  Information and Modeling}, vol.~61, no.~4, pp. 1560--1569, 2021.

\bibitem[Hochreiter and Schmidhuber(1997)]{LSTM:1997}
S.~Hochreiter and J.~Schmidhuber, ``Long short-term memory,'' \emph{Neural
  computation}, vol.~9, no.~8, pp. 1735--1780, 1997.

\bibitem[Loshchilov and Hutter(2017)]{AdamW:2017}
\BIBentryALTinterwordspacing
I.~Loshchilov and F.~Hutter, ``Decoupled weight decay regularization,'' 2017.
  [Online]. Available: \url{https://arxiv.org/abs/1711.05101}
\BIBentrySTDinterwordspacing

\bibitem[Bickerton et~al.(2012)Bickerton, Paolini, Besnard, Muresan, and
  Hopkins]{QEDscore:2012}
G.~R. Bickerton, G.~V. Paolini, J.~Besnard, S.~Muresan, and A.~L. Hopkins,
  ``Quantifying the chemical beauty of drugs,'' \emph{Nature chemistry},
  vol.~4, no.~2, pp. 90--98, 2012.

\bibitem[Ertl and Schuffenhauer(2009)]{SAS:2009}
P.~Ertl and A.~Schuffenhauer, ``Estimation of synthetic accessibility score of
  drug-like molecules based on molecular complexity and fragment
  contributions,'' \emph{Journal of cheminformatics}, vol.~1, no.~1, pp. 1--11,
  2009.

\end{thebibliography}
\bibliographystyle{IEEEtranN}
}

\end{document}